\documentclass[11pt,letterpaper]{article}
\usepackage{naaclhlt2016}
\usepackage{times}
\usepackage{latexsym}

\naaclfinalcopy 
\def\naaclpaperid{***} 

\usepackage{amsfonts}
\usepackage{mathrsfs}
\usepackage{amsmath}
\usepackage{amssymb}
\DeclareMathOperator*{\argmax}{argmax}
\usepackage{graphicx}
\usepackage{eqnarray}
\usepackage{multirow}
\usepackage{color}
\usepackage[ruled,longend]{algorithm2e}
\newcommand{\aesi}{\textit{AESI}}

\title{Joint Learning Templates and Slots for Event Schema Induction}

\author{Lei Sha, Sujian Li, Baobao Chang, Zhifang Sui\\
	    Key Laboratory of Computational Linguistics, Ministry of Education\\
	    School of Electronics Engineering and Computer Science, Peking University\\
	    Collaborative Innovation Center for Language Ability, Xuzhou 221009 China\\
	    {\tt {shalei, lisujian, chbb, szf}@pku.edu.cn}
	 }

\date{}

\begin{document}
\maketitle
\begin{abstract}
Automatic event schema induction (\aesi) means to extract meta-event from raw text, in other words, to find out what types (templates) of event may exist in the raw text and what roles (slots) may exist in each event type. In this paper, we propose a joint entity-driven model to learn templates and slots simultaneously based on the constraints of templates and slots in the same sentence. In addition, the entities' semantic information is also considered for the inner connectivity of the entities. We borrow the \textit{normalized cut} criteria in image segmentation to divide the entities into more accurate template clusters and slot clusters.  The experiment shows that our model gains a relatively higher result than  previous work.
\end{abstract}
\section{Introduction}\label{intro}
Event schema is a high-level representation of a bunch of similar events. It is very useful for the traditional information extraction (\textit{IE})\cite{sagayam2012survey} task. An example of event schema is shown in Table~\ref{mainslots}. Given the bombing schema, we only need to find proper words to fill the slots when  extracting a bombing event.

There are two main approaches for \aesi\ task. Both of them use the idea of clustering the potential event arguments to find the event schema. One of them is probabilistic graphical model \cite{joint,cheung2013probabilistic}. By incorporating templates and slots as latent topics, probabilistic graphical  models learns those templates and slots that best explains the text.  However, the graphical models considers the entities independently and do not take the interrelationship between entities into account.  Another method relies on ad-hoc clustering algorithms \cite{filatova2006automatic,sekine2006demand,pipeline}. \cite{pipeline} is a pipelined approach. In the first step, it uses pointwise mutual information(PMI) between any two clauses in the same document to learn events, and then learns syntactic patterns as fillers. However, the pipelined approach suffers from the error propagation problem, which means the errors in the template clustering can lead to more errors in the slot clustering.

\begin{table}
\begin{center}
\textbf{Bombing Template}


\begin{tabular}{ll}
  \textit{Perpetrator:} & person \\
  \textit{Victim:} & person \\
  \textit{Target:} & public \\
  \textit{Instrument:} & bomb \\
\end{tabular}
\end{center}
  \caption{The event schema of bombing event in MUC-4, it has a bombing template and four main slots}\label{mainslots}
\end{table}

This paper proposes an entity-driven model which jointly learns templates and slots for event schema induction. The main contribution of this paper are as follows:

\begin{itemize}
\item To better model the inner connectivity between entities, we borrow the  normalized cut in image segmentation as the clustering criteria.
\item  We use constraints between templates and between slots in one sentence to improve \aesi\ result.
\end{itemize}



\section{Task Definition}
\begin{figure}
  \centering
  \includegraphics[width=7.5cm]{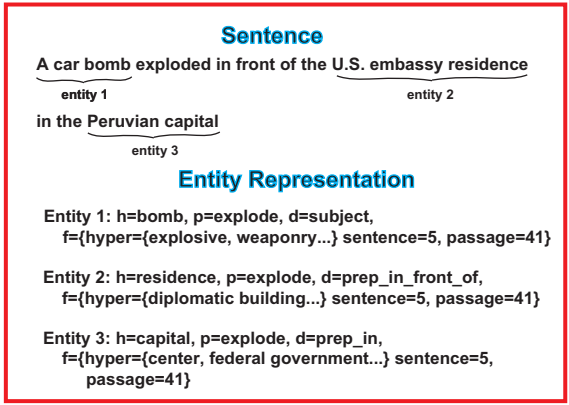}\\
  \caption{An entity example}\label{entity}
\end{figure}

Our model is an entity-driven model. This model represents a document $d$ as a series of entities $E_d=\{e_i|i=1,2,\cdots\}$. Each entity is a quadruple $e=(h,p,d,f)$. Here, $h$ represents the head word of an entity, $p$ represents its predicate, and $d$ represents the dependency path between the predicate and the head word, $f$ contains the features of the entity (such as the \textbf{direct} hypernyms of the head word), the sentence id where $e$ occurred and the document id where $e$ occurred. A simple example is Fig~\ref{entity}.


Our ultimate goal is to assign two labels, a slot variable $s$ and a template variable $t$, to each entity. After that, we can summarize all of them to get event schemas.

\section{Automatic Event Schema Induction}

\subsection{Inner Connectivity Between Entities}
We focus on two types of inner connectivity: (1) the likelihood of two entities to belong to the same template; (2) the likelihood of two entities to belong to the same slot;

\subsubsection{Template Level Connectivity}
It is easy to understand that entities occurred near each other are more likely to belong to  the same template. Therefore, \cite{pipeline} uses PMI to measure the correlation of two words in the same document, but it cannot put two words from different documents together. In the Bayesian model of \cite{joint}, p(predicate) is the key factor to decide the template, but it ignores the fact that entities occurring nearby should belong to the same template.  In this paper, we try to put two measures together. That is, if two entities occurred nearby, they can belong to the same template; if they have  similar meaning, they can also belong to the same template. We use PMI to measure the distance similarity and use word vector \cite{mikolov2013efficient} to calculate the semantic similarity.

 A word vector can well represent the meaning of a word. So we concatenate the word vector of the $j$-th entity's head word and its predicate, denoted as $vec_{hp}(i)$. We use the cosine distance $\cos_{hp}(i,j)$ to measure the difference of two vectors.

Then we can get the template level connectivity formula as shown in Eq~\ref{templatesimilarity}. The $PMI(i,j)$ is calculated by the head words of  entity mention $i$ and $j$.
\begin{equation}\label{templatesimilarity}
  \begin{aligned}
  W_T(i,j)=PMI(i,j)+\cos_{hp}(i,j)\\
\end{aligned}
\end{equation}
\subsubsection{Slot Level Connectivity}
If two entities can play  similar role in an event, they are likely to fill the same slot. We know that if two entities can play  similar role, their head words may have the same hypernyms. We only consider the \textbf{direct} hypernyms here. Also, their predicates may have similar meaning and the entities may have the same dependency path to their predicate. Therefore, we give the factors equal weights and add them together to get the slot level similarity.
\begin{equation}\label{slotsimilarity}
 \begin{aligned}
   W_S(i,j)&=\cos_p(i,j)+\delta(depend_i=depend_j)\\
&+\delta(hypernym_i\cap hypernym_j\ne \phi)
  \end{aligned}
\end{equation}
Here, the $\delta(\cdot)$ has value 1 when the inner expression is true and 0 otherwise. The ``hypernym" is derived from  Wordnet\cite{miller1995wordnet}, so it is a set of direct hypernyms. If two entities' head words have at least one common direct hypernym, then they may belong to the same slot. And again $\cos_p(i,j)$ represents the cosine distance between the predicates' word vector of entity $i$ and entity $j$.
\subsection{Template and Slot Clustering Using Normalized Cut}
 Normalized cut intend to maximize the intra-class similarity while minimize the inter class similarity, which deals well with the connectivity between entities.

We represent  each entity as a point in a high-dimension space. The edge weight between two points is their template level similarity / slot level similarity. Then the larger the similarity value is, the more likely the two entities (point) belong to the same template / slot, which is also our basis intuition.

For simplicity, denote the entity set as $E=\{e_1,\cdots,e_{|E|}\}$, and the  template set as $T$. We use the $|E|\times |T|$ partition matrix $X_T$ to represent the template clustering result. Let $X_T=[X_{T_1},\cdots,X_{T_{|T|}}]$, where $X_{T_l}$ is a binary indicator for template $l$($T_l$).
\begin{equation}
  X_T(i,l)=\left\{
        \begin{aligned}
           1 & ~~~e_i\in T_l \\
           0 & ~~~otherwise
        \end{aligned} \right.
\end{equation}
Usually, we define the degree matrix $D_T$ as: $D_T(i,i)=\!\!\sum_{j\in E}W_T(i,j), i=1,\cdots,|E|$. Obviously, $D_T$ is a diagonal matrix. It contains information about the weight sum of edges attached to each vertex. Then  we have the template clustering optimization  as shown in Eq~\ref{tobj} according to \cite{shi2000normalized}.
\begin{equation}\label{tobj}
\begin{aligned}
  \max\quad &\varepsilon_1(X_T)=\frac{1}{|T|}\sum^{|T|}_{l=1}\frac{X_{T_l}^TW_TX_{T_l}}{X_{T_l}^TD_TX_{T_l}}\\
  s.t.\quad &X_T\in\{0,1\}^{|E|\times |T|}~~~X_T\textbf{1}_{|T|}=\textbf{1}_{|E|}
\end{aligned}
\end{equation}
where $\textbf{1}_{|E|}$ represents the $|E|\times 1$ vector of all 1's.

For the slot clustering, we have a similar optimization  as shown in Eq~\ref{sobj}.
\begin{equation}\label{sobj}
\begin{aligned}
  \max\quad &\varepsilon_2(X_S)=\frac{1}{|S|}\sum^{|S|}_{l=1}\frac{X_{S_l}^TW_SX_{S_l}}{X_{S_l}^TD_SX_{S_l}}\\
  s.t.\quad &X_S\in\{0,1\}^{|E|\times |S|}~~~X_S\textbf{1}_{|S|}=\textbf{1}_{|E|}
\end{aligned}
\end{equation}
where $S$ represents the slot set, $X_S$ is the slot clustering result with $X_S=[X_{S_1},\cdots,X_{S_{|S|}}]$, where $X_{S_l}$ is a binary indicator for slot $l$($S_l$).
\begin{equation}
  X_S(i,l)=\left\{
        \begin{aligned}
           1 & ~~~e_i\in S_l \\
           0 & ~~~otherwise
        \end{aligned} \right.
\end{equation}
\subsection{Joint  Model With Sentence Constraints}
For event schema induction, we find an important property and we name it  ``Sentence constraint". The entities in one sentence often belong to one template but different slots.

The sentence constraint contains two types of constraint, ``template constraint" and ``slot constraint".
 \begin{enumerate}
   \item \textbf{Template constraint}: Entities in the same sentence are usually in the same template. Hence we should make the templates taken by a sentence as few as possible.
   \item \textbf{Slot constraint}: Entities in the same sentence are usually in different slots. Hence we should make the slots taken by a sentence as many as possible.
 \end{enumerate}
Based on these consideration, we can add an extra item to the optimization object. Let $N_{sentence}$ be the number of sentences. Define $N_{sentence}\times |E|$ matrix $J$ as the sentence constraint matrix, the entries of $J$ is as following:
\begin{equation}\label{J}
  J(i,j)=\left\{
        \begin{aligned}
           1 & ~~~e_i\in Sentence_j \\
           0 & ~~~otherwise
        \end{aligned} \right.
\end{equation}
Easy to show, the product $G_T=J^TX_T$ represents the relation between sentences and templates. In matrix $G_T$, the $(i,j)$-th entry represents how many entities in sentence $i$ are belong to $T_j$.

Using $G_T$, we can construct our objective. To represent the two constraints, the best objective we have found is the trace value: $tr(G_TG_T^T)$. Each entry on the diagonal of matrix $G_TG_T^T$ is the square sum of all the entries in the corresponding line in $G_T$, and the larger the trace value is, the less templates the sentence would taken. Since $tr(G_TG_T^T)$ is the sum of the diagonal elements, we only need to maximize the value $tr(G_TG_T^T)$ to meet the template constraint. For the same reason, we need to minimize the value $tr(G_SG_S^T)$ to meet the slot constraint.

Generally, we have the following optimization objective:
\begin{equation}\label{Jobj}
  \varepsilon_3(X_T,X_S)=\frac{tr\big(X_T^TJJ^TX_T\big)}{tr\big(X_S^TJJ^TX_S\big)}
\end{equation}
The whole joint model is shown in Eq~\ref{totalobj}. The solving method is in the attachment file.
\begin{equation}\label{totalobj}
 \begin{aligned}
  X_T,X_S&=\mathop{\argmax}_{X_T,X_S}\varepsilon_1(X_T)+\varepsilon_2(X_S)+\varepsilon_3(X_T,X_S)\\
  s.t.&~ X_T\in\{0,1\}^{|E|\times |T|}~~~X_T\textbf{1}_{|T|}=\textbf{1}_{|E|}\\
   &~X_S\in\{0,1\}^{|E|\times |S|}~~~ X_S\textbf{1}_{|S|}=\textbf{1}_{|E|}
 \end{aligned}
\end{equation}
\section{Experiment}
\subsection{Dataset}
In this paper, we use MUC-4\cite{sundheim1991third} as our dataset, which is the same as previous works \cite{pipeline,joint}. MUC-4 corpus contains 1300 documents in the training set, 200 in development set (TS1, TS2) and 200 in testing set (TS3, TS4) about Latin American news of terrorism events. We ran several times on the 1500 documents (training/dev set) and choose the best $|T|$ and $|S|$ as $|T| = 6$, $|S| = 4$. Then we report the performance  of  test set. For each document, it provides a series of hand-constructed event schemas, which are called  gold schemas.  With these gold schemas we can evaluate our results. The MUC-4 corpus contains six template types: \textbf{Attack, Kidnapping, Bombing, Arson, Robbery, and Forced Work Stoppage}, and for each template, there are 25 slots. Since most previous works do not evaluate their performance on all the 25 slots, they instead focus on 4 main slots like Table~\ref{mainslots}, we will also focus on these four slots. We use the Stanford CoreNLP toolkit to parse the MUC-4 corpus.
\subsection{Performance}
\begin{figure}
  \centering
  \includegraphics[width=8cm]{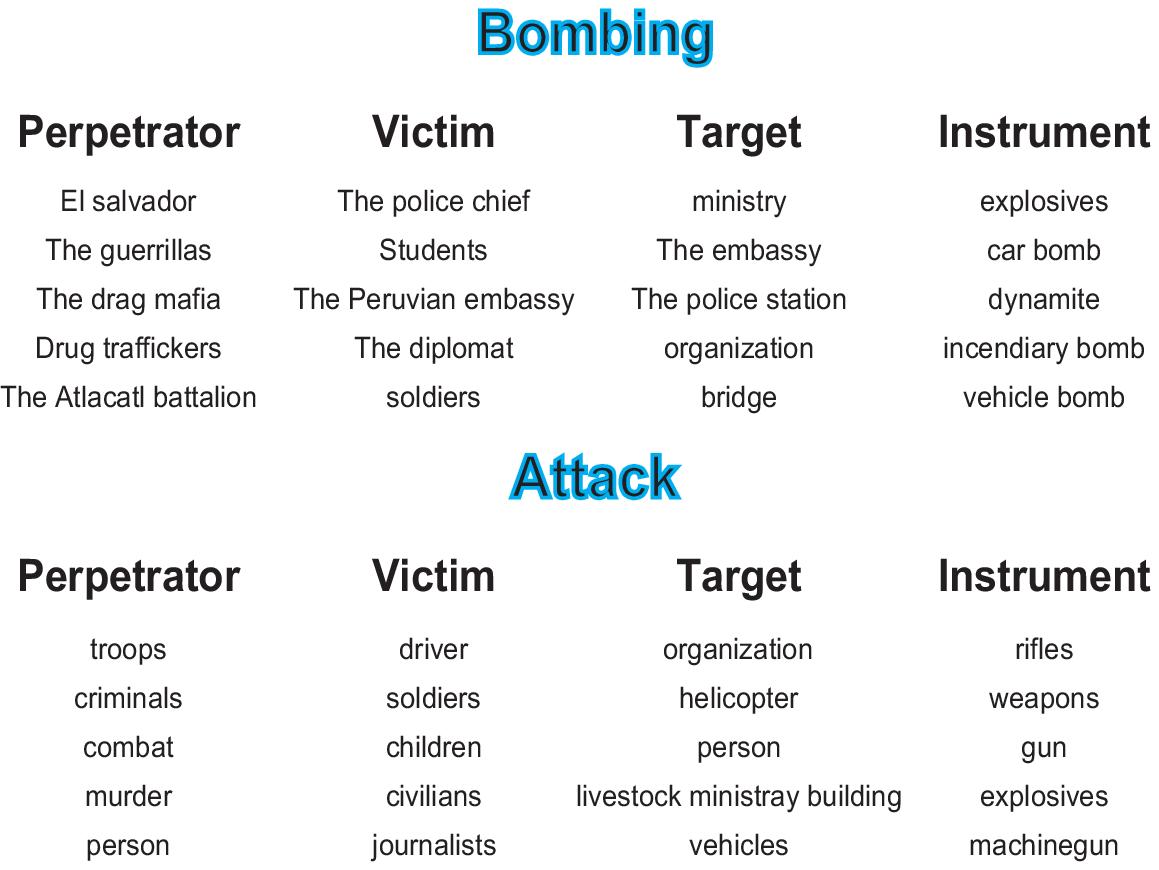}\\
  \caption{Part of the result}\label{res}
\end{figure}

Fig~\ref{res} shows two examples of our learned schemas: Bombing and Attacking. The five words in each slot are the five randomly picked entities from the mapped slots. The templates and slots that were joint learned seem reasonable.

We compare our results with four works \cite{pipeline,cheung2013probabilistic,joint,nguyen-EtAl:2015:ACL-IJCNLP1} as is shown in Table~\ref{Sresult}. Our model has outperformed all of the previous methods. The improvement of recall is due to the normalized cut criteria, which can better use   the inner connectivity between entities. The sentence constraint improves the result one step further.
\begin{table}
  \centering
  \begin{tabular}{lccc}
  \hline
                     & Prec & Recall& F1 \\\hline
  C\&J~(2011)      & \textbf{0.48} & 0.25 & 0.33 \\
  \newcite{cheung2013probabilistic}  & 0.32 & 0.37 & 0.34 \\
  \newcite{joint}    & 0.41 & 0.41 & 0.41 \\
  \newcite{nguyen-EtAl:2015:ACL-IJCNLP1}    & 0.36 & 0.54 & 0.43 \\\hline
  Our Model-SC    & 0.38 & 0.68 & 0.49 \\
  Our Model    & 0.39 & \textbf{0.70} & \textbf{0.50} \\
  \hline
\end{tabular}
  \caption{Comparison to state-of-the-art unsupervised systems, ``-SC'' means without sentence constraint}\label{Sresult}
\end{table}
\section{Related Works}\label{relatedwork}
\aesi\ task has been researched for many years. \newcite{shinyama2006preemptive} proposed an approach to learn templates with unlabeled corpus. They use \emph{unrestricted relation discovery} to discover relations in unlabeled corpus as well as extract their fillers. Their constraints are that they need redundant documents and their relations are binary over repeated named entities. \cite{chen2011domain} also extract binary relations using generative model.

\newcite{kasch2010mining}, \newcite{chambers2008unsupervised}, \newcite{chambers2009unsupervised}, \newcite{balasubramanian2013generating} captures template-like knowledge from unlabeled text by large-scale learning of scripts and narrative schemas. However, their structures (template/slot) are limited to frequent topics in a large corpus. \newcite{pipeline} uses their idea, and their goal is to characterize a specific domain with limited data using a three-stage clustering algorithm.

Also, there are some state-of-the-art works using probabilistic graphic model \cite{joint,cheung2013probabilistic,nguyen-EtAl:2015:ACL-IJCNLP1}.

\section{Conclusion}

This paper presented a  joint entity-driven model  to induct event schemas automatically.

This model uses word embedding as well as PMI to  measure the inner connection of  entities and uses normalized cut for more accurate clustering. Finally, our model uses sentence constraint to extract templates and slots simultaneously. The experiment has proved the effectiveness of our model.
\section*{Acknowledgments}
This research is supported by National Key Basic Research Program of China (No.2014CB340504) and National Natural Science Foundation of China (No.61375074,61273318). The contact authors of this paper are Sujian Li and Baobao Chang.

\bibliography{refs}
\bibliographystyle{naaclhlt2016}
\end{document}


\title{The Solving Method of the Event Schema Induction Joint Model}
\author{Lei Sha}
\maketitle
\section{The Model}
Template clustering optimization  is shown in Eq~\ref{tobj}.
\begin{equation}\label{tobj}
\begin{aligned}
  \max\quad &\varepsilon_1(X_T)=\frac{1}{|T|}\sum^{|T|}_{l=1}\frac{X_{T_l}^TW_TX_{T_l}}{X_{T_l}^TD_TX_{T_l}}\\
  s.t.\quad &X_T\in\{0,1\}^{|E|\times |T|}~~~X_T\textbf{1}_{|T|}=\textbf{1}_{|E|}
\end{aligned}
\end{equation}
Here, $\textbf{1}_{|E|}$ represents the $|E|\times 1$ vector of all 1's.

Slot clustering optimization   is shown in Eq~\ref{sobj}.
\begin{equation}\label{sobj}
\begin{aligned}
  \max\quad &\varepsilon_2(X_S)=\frac{1}{|S|}\sum^{|S|}_{l=1}\frac{X_{S_l}^TW_SX_{S_l}}{X_{S_l}^TD_SX_{S_l}}\\
  s.t.\quad &X_S\in\{0,1\}^{|E|\times |S|}~~~X_S\textbf{1}_{|S|}=\textbf{1}_{|E|}
\end{aligned}
\end{equation}
Here, $S$ represents the slot set, $X_S$ is the slot clustering result with $X_S=[X_{S_1},\cdots,X_{S_{|S|}}]$, where $X_{S_l}$ is a binary indicator for slot $l$($S_l$).
\begin{equation}
  X_S(i,l)=\left\{
        \begin{aligned}
           1 & ~~~e_i\in S_l \\
           0 & ~~~otherwise
        \end{aligned} \right.
\end{equation}

The original sentence constraint model is shown as follows:
\begin{equation}\label{Jobj1}
\begin{aligned}
  \varepsilon_3(X_T,X_S)=\frac{tr\big(X_T^TJJ^TX_T\big)}{tr\big(X_S^TJJ^TX_S\big)}\\
\end{aligned}
\end{equation}
However, this form of objective is hard to optimize, we can transfer the slot constraint objective $tr(G_SG_S^T)$ ($G_S=J^TX_S$) to something that should be maximized. Since $tr(G_SG_S^T) = tr\big(X_S^T JJ^TX_S)\big) $, to minimize $tr\big(X_S^T JJ^TX_S)\big) $ is the same as to maximize $tr\big(X_S^T(E-JJ^T)X_S)\big) $ ($E=\textbf{1}\cdot\textbf{1}^T$). $\textbf 1$ represents an all 1 vector. It can be proved that
$tr\big(X_S^T(E-JJ^T)X_S)\big) $ is positive.

Generally, we have the following optimization objective:
\begin{equation}\label{Jobj}
\begin{aligned}
  \max&~ \varepsilon_3(X_T,X_S)=tr\big(X_T^TJJ^TX_T\big)tr\big(X_S^T(E-JJ^T)X_S)\big)\\
  s.t.&~ X_T\in\{0,1\}^{|E|\times |T|}~~~X_T\textbf{1}_{|T|}=\textbf{1}_{|E|}\\
   &~X_S\in\{0,1\}^{|E|\times |S|}~~~ X_S\textbf{1}_{|S|}=\textbf{1}_{|E|}
\end{aligned}
\end{equation}
The whole joint model is shown in Eq~\ref{totalobj}.  The first item represents the goodness of the templates clustering. The second item represents the goodness of the slot clustering. The third item is the sentence constraint item. However, this model is too complex  to be solved by normal optimization method. Therefore, we use the Alternating Maximization Procedure\cite{naiss2010alternating} to solve this problem in the following section.
\begin{equation}\label{totalobj}
 \begin{aligned}
  X_T,X_S&=\mathop{\argmax}_{X_T,X_S}\varepsilon_1(X_T)+\varepsilon_2(X_S)+\varepsilon_3(X_T,X_S)\\
  s.t.&~ X_T\in\{0,1\}^{|E|\times |T|}~~~X_T\textbf{1}_{|T|}=\textbf{1}_{|E|}\\
   &~X_S\in\{0,1\}^{|E|\times |S|}~~~ X_S\textbf{1}_{|S|}=\textbf{1}_{|E|}
 \end{aligned}
\end{equation}
\section{Solving Method: Alternating Maximization Procedure(\textit{AMP}) }\label{solve}
In this section, the detailed solving method of the complex model shown in Eq~\ref{totalobj} will be  illustrated. The ultimate objective in Eq~\ref{totalobj} is the combination of optimization objective in Eq~\ref{tobj}, Eq~\ref{sobj} and Eq~\ref{Jobj}.

The first two items  in Eq~\ref{totalobj} is the form of generalized Rayleigh quotient and  can be solved using the method in \cite{yu2003multiclass}, which mainly contains two steps: 1) find the continuous optimal value 2) discretization. We use the   \textit{AMP} method to get the numerical solution of Eq~\ref{totalobj}. The \textit{AMP} algorithm can be viewed as a joint maximization method by fixing one argument and maximizing over the other. After we fixed $X_S$ or $X_T$, we can transform the objective to the form of generalized Rayleigh quotient which could be solved by the method in \cite{yu2003multiclass}.

\paragraph{When $X_T$ is fixed}
The first term in Eq~\ref{totalobj} is a constant in this case, so that we ignore it for simplicity. Let $f(X_T)=tr\big(X_T^T JJ^TX_T\big)$, then Eq~\ref{totalobj} becomes:
\begin{equation}\label{XT2}
 \max~\varepsilon(X_S;X_T)=\frac{1}{|S|}\sum^{|S|}_{l=1}\frac{X_{S_l}^TW_SX_{S_l}}{X_{S_l}^TD_SX_{S_l}}+f(X_T)\sum^{|S|}_{l=1}X_{S_l}^T(E-JJ^T)X_{S_l}
\end{equation}
We can  reduce the fractions to a common denominator, then Eq~\ref{XT2} becomes:
\begin{equation}\label{XT3}
\begin{aligned}
\sum^{|S|}_{l=1}\frac{\frac{1}{|S|}X_{S_l}^TW_SX_{S_l}+f(X_T)\ast X_{S_l}^T(E-JJ^T)X_{S_l}X_{S_l}^TD_SX_{S_l}}{X_{S_l}^TD_SX_{S_l}}\\
\end{aligned}
\end{equation}
Note that the term $X_{S_l}^T(E-JJ^T)X_{S_l}X_{S_l}^TD_SX_{S_l}$ is a scalar, so that we can take it as a trace of a $1\times 1$ matrix as shown in Eq~\ref{XT4}.
\begin{equation}\label{XT4}
\begin{aligned}
&X_{S_l}^T(E-JJ^T)X_{S_l}X_{S_l}^TD_SX_{S_l}\\
&=tr(X_{S_l}^T(E-JJ^T)X_{S_l}X_{S_l}^TD_SX_{S_l})\\
&=\Omega_S X_{S_l}^T(E-JJ^T)D_SX_{S_l}
\end{aligned}
\end{equation}
Here, $\Omega_S=X_{S_l}^T X_{S_l}$ is a diagonal matrix. Each diagonal entry is the number of entities in the corresponding slot.

In order to represent Eq~\ref{XT3} to the form of Eq~\ref{XT5}, we need to keep $D^\ast_S=D_S$, and the $W^\ast_S$ is as Eq~\ref{WS}. In order to keep $W^\ast_S$ a symmetric matrix, we add $\frac{1}{2}$ of Eq~\ref{XT4} to both sides of $X_{S_l}^TW_SX_{S_l}$.
\begin{equation}\label{XT5}
\begin{aligned}
\varepsilon(X_S;X_T)=\sum^{|S|}_{l=1}\frac{X_{S_l}^TW^\ast_SX_{S_l}}{X_{S_l}^TD^\ast_SX_{S_l}}\\
\end{aligned}
\end{equation}
\begin{equation}\label{WS}
\left\{\begin{aligned}
W^\ast_S&=\frac{1}{2}f(X_T)D_S(E-JJ^T)\Omega_S+\frac{1}{|S|}W_S\\
        &+\frac{1}{2}f(X_T)\Omega_S(E-JJ^T)D_S\\
D^\ast_S&=D_S
\end{aligned}\right.
\end{equation}
\paragraph{When $X_S$ is fixed}
Using the same method as the above, in order to get the form of Eq~\ref{XS1}, the value of $W_T^\ast$ and $D_T^\ast$ are calculated as Eq~\ref{WT}.
\begin{equation}\label{XS1}
\begin{aligned}
\varepsilon(X_T;X_S)=\sum^{|T|}_{l=1}\frac{X_{T_l}^TW^\ast_TX_{T_l}}{X_{T_l}^TD^\ast_TX_{T_l}}\\
\end{aligned}
\end{equation}
\begin{equation}\label{WT}
\left\{
\begin{aligned}
W^\ast_T=&\frac{1}{2f(X_S)}JJ^T D_T\Omega_T+\frac{1}{|T|}W_T\\
        +&\frac{1}{2f(X_S)}\Omega_TD_TJJ^T\\
D^\ast_T=&D_T\\
\end{aligned}
\right.
\end{equation}

\paragraph{Stopping criteria}
According to \cite{yu2003multiclass}, if $X_T, X_S$ is a feasible solution to Eq~\ref{totalobj}, so is $\{X_TR_T,X_SR_S|R_T^TR_T=I,R_S^TR_S=I\}$, and they have the same objective value: $\varepsilon(X_TR_T,X_SR_S)=\varepsilon(X_T,X_S)$. Therefore, if Eq~\ref{stop} is satisfied, the loop ends.

\begin{equation}\label{stop}
  \begin{aligned}
  &\|X_T^{new}-X_T^{old}R_T\|=0\\
  &\|X_S^{new}-X_S^{old}R_S\|=0\\
  \end{aligned}
\end{equation}
We can get the closed form of $R_T$ and $R_S$ as shown in Eq~\ref{R}.
\begin{equation}\label{R}
  \begin{aligned}
  &R_T=(X_T^{(new)T}X_T^{new})^{-1}X_T^{(new)T}X_T^{old}\\
  &R_S=(X_S^{(new)T}X_S^{new})^{-1}X_S^{(new)T}X_S^{old}\\
  \end{aligned}
\end{equation}
Therefore, the ultimate stop criteria becomes $\|R_T^TR_T-I\|+\|R_S^TR_S-I\|<\epsilon$, $\epsilon$ is very close to 0.

The total algorithm of the whole process is shown as Algorithm~\ref{algorithm}.
Since the optimization objective is a differentiable function, the convergence to the optimum solution can be guaranteed by \cite{naiss2010alternating,hastie2009elements}.

\begin{algorithm}\label{algorithm}
  \caption{The pseudo code of the optimum value finding process}
     \KwIn{\\
      \Indp Template level similarity matrix, $W_T$\;
      Slot level similarity matrix, $W_S$\;
      sentence constraint matrix, $J$.}
     \KwOut{\\
       \Indp The partition matrix of template, $X_T$\;
      The partition matrix of slot, $X_S$\;}

   \Begin{
      Randomly initialize $X_T$ and $X_S$\;
    \While {$\|R_T^TR_T-I\|+\|R_S^TR_S-I\|>\epsilon$}
    {
       Fix $X_T$, calculate Eq~\ref{WS}\;
       Find $X_S$ which can maximize Eq~\ref{XT5}\;
       Fix $X_S$, calculate Eq~\ref{WT}\;
       Find $X_T$ which can maximize Eq~\ref{XS1}\;
       Calculate $R_T$ and $R_S$ by Eq~\ref{R}\;
    }
    Discretize $X_T$ and $X_S$\;
    \Return{ $X_T$ and $X_S$}
  }
\end{algorithm}
\section{Experiment Setting}
The $\Omega_T$ and $\Omega_S$ in Eq~\ref{WT} and Eq~\ref{WS} can be seen as a prior of the template cluster size and slot cluster size. We use the most na\"{i}ve prior that all clusters are of the same size.
\bibliography{refs}
\bibliographystyle{plain}